\documentclass{article}
\usepackage{arxiv}

\usepackage{amssymb,amsthm,amsmath}
\usepackage{xcolor,paralist,hyperref,titlesec,fancyhdr,etoolbox}

\usepackage[T1]{fontenc}
\usepackage{graphicx}
\usepackage{booktabs}
\usepackage{amsmath}
\usepackage{amssymb}
\usepackage{multirow}
\usepackage{xcolor}
\usepackage{hyperref}
\usepackage{natbib}
\usepackage{tikz}
\usetikzlibrary{positioning, arrows.meta, calc}
\usepackage[acronym,nonumberlist,nogroupskip,nohypertypes={acronym}]{glossaries}
\makenoidxglossaries

\newacronym{qd}{QD}{Quality-Diversity}
\newacronym{sail}{SAIL}{Surrogate-Assisted Illumination Algorithm}
\newacronym{ucb}{UCB}{Upper Confidence Bound}
\newacronym{gp}{GP}{Gaussian Process}
\newacronym{svgp}{SVGP}{Sparse Variational Gaussian Process}
\newacronym{cfd}{CFD}{Computational Fluid Dynamics}
\newacronym{dwd}{DWD}{Deutscher Wetterdienst}
\newacronym{ard}{ARD}{Automatic Relevance Determination}
\newacronym{hpc}{HPC}{High-Performance Computing}
\newacronym{grz}{GRZ}{site coverage ratio, \emph{Grundfl\"achenzahl}}
\newacronym{gfz}{GFZ}{floor area ratio, \emph{Geschossfl\"achenzahl}}
\newacronym{baunvo}{BauNVO}{\emph{Baunutzungsverordnung}, German Building Use Ordinance}

\titlespacing*{\section}{0pt}{0ex}{0ex}

\hypersetup{ colorlinks=true, linkcolor=black, filecolor=black, urlcolor=black }

\usepackage{lipsum}

\begin{document}
\title{U-Net-Accelerated Quality-Diversity Optimization for Climate-Adaptive Urban Layouts}

\author{ 
\href{https://orcid.org/0000-0002-8668-1796}{\includegraphics[scale=0.06]{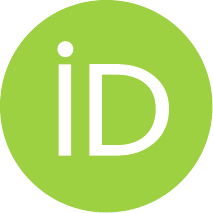}\hspace{1mm}Alexander~Hagg}\\
Institute of Technology, Resource and Energy-efficient Engineering (TREE) \\
Bonn-Rhein-Sieg University of Applied Sciences \\
Sankt Augustin, 53757, Germany \\
\texttt{alexander.hagg@h-brs.de} \\
\And
\href{https://orcid.org/0000-0001-8377-9053}{\includegraphics[scale=0.06]{orcid.pdf}\hspace{1mm}Tania~Guerrero}\\
Institute of Technology, Resource and Energy-efficient Engineering (TREE) \\
Bonn-Rhein-Sieg University of Applied Sciences \\
Sankt Augustin, 53757, Germany \\
\texttt{tania.guerrero@smail.inf.h-brs.de} \\
\And
\href{https://orcid.org/0000-0003-1480-6745}{\includegraphics[scale=0.06]{orcid.pdf}\hspace{1mm}Dirk~Reith}\\
Institute of Technology, Resource and Energy-efficient Engineering (TREE) \\
Bonn-Rhein-Sieg University of Applied Sciences \\
Sankt Augustin, 53757, Germany \\
Fraunhofer Institute for Algorithms and Scientific Computing (SCAI), Sankt Augustin, Germany\\
\texttt{dirk.reith@h-brs.de} \\
}

\maketitle


\begin{abstract}
Optimizing urban layouts for climate adaptation requires balancing building density with cold-air ventilation. Because physics-based climate simulations are computationally expensive, planners typically evaluate fewer than ten manual designs. \gls{qd} algorithms offer a way to systematically illuminate the design space, but they require surrogate models to be practical.

In this paper, we replace a slow, regulatory physics simulator with a spatial deep-learning surrogate (U-Net) inside an offline MAP-Elites loop. We systematically compare this spatial approach with a traditional \gls{gp} surrogate across different training-data strategies (quasi-random Sobol sampling vs.\ active \gls{qd} bootstrapping).

Our results reveal that scalar \gls{gp} surrogates fail catastrophically when trained on random samples, requiring expensive, actively generated \gls{qd} archives to generalize. In contrast, the spatial inductive bias of the U-Net allows it to learn the underlying physics mapping robustly ($R^2 = 0.996$), completely independent of the training data source. This allows offline \gls{qd} optimization to achieve highly accurate fitness rankings ($\rho = 0.994$) using only a one-time batch of random training samples. The resulting pipeline, deployed in the open-source OpenSKIZZE tool, generates thousands of diverse, climate-evaluated building layouts in under ten minutes.
\end{abstract} 

\keywords{Quality-Diversity Optimization, Surrogate-Assisted Optimization, Urban Climate Modeling, MAP-Elites, Deep Learning Surrogates, Urban Planning}

\section{Introduction}
\label{sec:intro}

Urban densification -- the process of increasing building density within existing settlement boundaries -- is a central strategy for reducing land consumption and supporting sustainable urbanization.
However, denser urban fabric can obstruct nocturnal cold air drainage flows that are essential for mitigating urban heat islands~\cite{alcoforado2009application,Kruse2018lanuv}.
Cold air forms at night over vegetated and open areas, and gravity-driven katabatic flows transport this cooler air into adjacent built-up zones, providing natural ventilation.
New construction that blocks these corridors can measurably increase heat stress for residents and degrade urban climate resilience~\cite{berchtold2023MUTABOR}.

Physics-based simulation tools such as the KLAM\_21 cold air drainage model~\cite{KLAM21} -- the German regulatory standard used by the national weather service (\gls{dwd}) -- can quantify these effects, but a single evaluation takes 5--20 minutes for isolated parcels and up to 24 hours for realistic city-scale domains.
In practice, urban planners manually design and compare fewer than ten building layout variants, making systematic exploration of the design space effectively impossible.
\Gls{qd} algorithms~\cite{Mouret2015,Chatzilygeroudis2021} such as MAP-Elites offer an alternative: rather than converging to a single optimum, they illuminate the full domain of possible designs~\cite{hagg2025full} by maintaining a diverse archive of high-performing solutions across multiple user-defined feature dimensions (e.g., site coverage, building height, open space proportion).
However, \gls{qd} algorithms commonly require $10^5$ to $10^6$ fitness evaluations, rendering direct coupling with physics simulators infeasible.

\paragraph{Contribution.}
We bridge this gap with a deep-learning surrogate that replaces the physics simulator entirely during \gls{qd} optimization (Fig.~\ref{fig:pipeline}).
Our systematic evaluation of training data strategies, surrogate architectures, and \gls{qd} configurations yields three actionable insights:

\noindent\textbf{(1)~\Gls{qd} bootstrapping is unnecessary for spatial surrogates.}
We compare data from \gls{sail}~\cite{Gaier2018DataEfficient} (\gls{qd} archives with real KLAM\_21 evaluations, which we initially hypothesized to be essential) with Sobol quasi-random samples evaluated by KLAM\_21.
For scalar \gls{gp} surrogates, \gls{qd}-generated data is indeed critical.
For the U-Net, however, Sobol data yields equivalent accuracy ($R^2 \geq 0.994$) -- simplifying the practical pipeline.

\noindent\textbf{(2)~Spatial U-Net surrogate.}
A U-Net~\cite{Ronneberger2015} trained on Sobol-sampled KLAM\_21 simulations predicts the full 6-channel spatial output from 3-channel input grids at $R^2 = 0.996$ and $\sim$2\,ms per evaluation.
The one-time training data cost ($\sim$19{,}000 simulations, trivially parallelizable on \gls{hpc}) is amortized across unlimited subsequent \gls{qd} runs -- each producing $\sim$100{,}000 surrogate evaluations in minutes rather than the $\sim$10{,}000 hours required by direct simulation.

\noindent\textbf{(3)~Offline \gls{qd} without exploration.}
MAP-Elites with the frozen U-Net produces diverse building layouts in 3--8 minutes.
Validation against KLAM\_21 ground truth yields Spearman $\rho = 0.994$ \emph{without} any uncertainty-guided exploration mechanism.

\noindent
Along the pipeline, four experiments address: 
\textbf{RQ1:} What training data strategy yields the best surrogate generalization, and does it depend on architecture?
\textbf{RQ2:} How do \gls{svgp} and U-Net compare for offline \gls{qd}?
\textbf{RQ3:} Can offline surrogate-only \gls{qd} reliably reproduce physics-validated archives?

As a proof of concept, we use isolated square parcels under simplified boundary conditions.
This setup isolates morphological effects on cold-air flux. In real urban contexts with irregular geometries, heterogeneous land use, variable winds, and multi-scale domains, surrogate accuracy and ranking fidelity will likely degrade; thus, our metrics represent upper bounds for real-world deployment. The full pipeline is deployed in OpenSKIZZE, an open-source web application for climate-adaptive urban design.

\section{Related Work}
\label{sec:related}

\paragraph{\Gls{qd} optimization.}
MAP-Elites~\cite{Mouret2015} partitions a user-defined feature space into cells and fills each with the highest-performing solution found.
Extensions include CVT-MAP-Elites~\cite{Vassiliades2018}, differentiable \gls{qd}~\cite{Fontaine2021}, and covariance matrix adaptation variants~\cite{Fontaine2020CMAME}.
The PyRibs framework~\cite{Tjanardjo2022} provides modular archive and emitter components used in this work.

\paragraph{Surrogate-assisted \gls{qd}.}
\gls{sail}~\cite{Gaier2018DataEfficient} introduced \gls{gp}-assisted MAP-Elites, using an acquisition function (\gls{ucb}) to balance fitness exploitation with model uncertainty exploration.
Subsequent work studied surrogate-assisted illumination in robotics~\cite{Keller2020} and aerodynamics~\cite{Gaier2018DataEfficient}, while Hagg et al.~\cite{hagg2020designing} showed that surrogate models can also replace the archive's features.
More recently, deep neural networks have replaced \glspl{gp} as state-of-the-art \gls{qd} surrogates. For instance, DSA-ME~\cite{Zhang2022DSAME} trains a DNN to predict objectives and measures in MAP-Elites for game design jointly, and DSAGE~\cite{Bhatt2022DSAGE} utilizes deep surrogates for environment generation. 
Additionally, Kent et al.~\cite{Kent2024BOQD} proposed Bayesian optimisation with coupled descriptor functions, and Flageat and Cully~\cite{Flageat2023UQD} addressed uncertainty quantification in noisy \gls{qd} domains.
A critical limitation of these deep \gls{qd} approaches, however, is that they map genomes directly to \emph{scalar} outputs (fitness or behavioral descriptors). Because they predict scalars, they remain highly susceptible to developing spurious optima in unseen regions of the input space~\cite{Gaier2018DataEfficient}. Consequently, they still require expensive active-learning loops -- interleaving surrogate predictions with true physics evaluations -- to correct these errors. 

More broadly, surrogate-assisted evolutionary computation traditionally balances evaluation cost and accuracy using scalar models (e.g., Kriging/GP~\cite{d1951statistical,williams2006gaussian}, Random Forests \cite{breiman2001random}), which struggle with high-dimensional spatial data. We replace scalar approximation with field-to-field spatial U-Net prediction. Modeling the complete physics mapping introduces a strong physical inductive bias that inherently resists spurious optima, enabling accurate, purely offline \gls{qd} optimization without active-learning loops.

\paragraph{Deep learning surrogates for physics.}
U-Net architectures~\cite{Ronneberger2015} have been applied as surrogates
for \gls{cfd}~\cite{Thuerey2020,Kochkov2021}, weather prediction~\cite{Pathak2022FourCastNet}, and urban microclimate modeling~\cite{Mortezazadeh2022}.
Conditional GANs~\cite{Isola2017} and adversarial image-to-image models~\cite{MillaVal2024} offer alternatives but require more careful training.
While architectures such as Fourier Neural Operators~\cite{Pathak2022FourCastNet} or conditional GANs offer promising alternatives for fluid dynamics, an exhaustive benchmarking of deep learning architectures is outside the scope of this work. Our primary objective is to demonstrate the fundamental advantage of the spatial representation paradigm over scalar genome-based approximation in QD optimization, leaving the exploration of more advanced spatial architectures for future studies.
For urban wind specifically, recent work has applied U-Nets to predict pedestrian-level wind distribution around buildings~\cite{Wang2025Wind}, to optimise city structures for wind resilience~\cite{Nowak2024UNetCity}, and to predict urban canopy flows from LES data~\cite{Vargiemezis2025UNetLES}.
However, none of these combine spatial surrogates with \gls{qd} optimization.
Our contribution is the first to leverage deep spatial surrogates within a \gls{qd} pipeline, enabling offline MAP-Elites that avoid the spurious-optima failure mode inherent to genome-based scalar surrogates.

\paragraph{Computational urban climate modeling.}
Simulating urban cold air flows ranges from full
\gls{cfd}~\cite{allegrini2015coupled} (hours per run) to simplified drainage
models such as KLAM\_21~\cite{KLAM21} (minutes).
Surrogate-based \gls{cfd} acceleration has been
explored~\cite{Papadopoulos2018,Zhang2023DLUrban}, but not for regulatory cold
air models combined with \gls{qd} optimization.

\section{Problem Formulation}
\label{sec:problem}

KLAM\_21~\cite{KLAM21} is a 2.5D katabatic cold air drainage model maintained by the \gls{dwd}.
Given terrain elevation, building heights, and land-use classification on a regular grid, it simulates nocturnal cold-air production, accumulation, and gravity-driven flow over multiple time steps.
The key output fields at pedestrian level (2\,m height) are the cold air content $E_x$ (in 100\,J/m$^2$) and horizontal wind velocity components $u_q, v_q$ (in m/s).

We define the optimization objective as the mean cold air energy flux over the region of interest:
\begin{equation}
\label{eq:flux}
    \Phi = \bar{E}_x \cdot \bar{u}_{2\text{m}}, \quad
    \bar{u}_{2\text{m}} = \sqrt{\bar{u}_q^2 + \bar{v}_q^2}
\end{equation}
where $\bar{E}_x$ and $\bar{u}_{2\text{m}}$ are spatial averages over the building parcel.
Higher $\Phi$ indicates better cold air ventilation.
The objective is to \emph{maximize} $\Phi$ while exploring diverse building configurations.
Prior analysis of the KLAM\_21 model~\cite{KLAM21,Kruse2018lanuv} indicates that \gls{grz} is the dominant morphological driver of pedestrian-level cold air flux, because cold air flows predominantly \emph{around} buildings at ground level rather than over them.

Each building layout is encoded as a 60-dimensional genome representing 10 rectangular buildings, each with 6 parameters: position ($x$, $y$), dimensions (width, length), height (in floors), and rotation angle.
Buildings can overlap, merging into complex footprints.
The genome is decoded into a 2D heightmap on a regular grid at 3\,m resolution, which serves as input to both physics simulation and surrogates (Fig.~\ref{fig:pipeline}).

The \gls{qd} archive uses 8 feature dimensions derived from planning practice:
(1)~\gls{grz}, (2)~\gls{gfz}, (3)~average building height, (4)~height variability, (5)~average building distance, (6)~building count, (7)~compactness, and (8)~park factor (open space proportion).
These correspond to established urban planning metrics under the \gls{baunvo} and enable planners to navigate the solution space using familiar descriptors.
We use a PyRibs~\cite{Tjanardjo2022} \texttt{GridArchive} with discretized bins for each feature dimension.

\begin{figure}[t]
\centering
\begin{tikzpicture}[>=Stealth, font=\footnotesize, scale=0.75, every node/.style={scale=0.9}]
  \fill[green!30] (0,0) rectangle (8.9,5);
  \fill[orange!15] (4.6,0) rectangle (8.9,5);
  \fill[white] (4.6,1.5) rectangle (6.6,3.5);
  \draw[very thick, red!80!black] (4.6,1.5) rectangle (6.6,3.5);
  \fill[gray!60] (4.8,1.7) rectangle (5.4,2.3);
  \fill[gray!60] (5.6,2.0) rectangle (6.2,2.9);
  \fill[gray!60] (5.0,2.8) rectangle (5.5,3.3);
  \draw[thick] (0,0) rectangle (8.9,5);
  \draw[->, very thick, blue!60] (0.5,2.5) -- (4.0,2.5);
  \node[blue!60, above, font=\footnotesize] at (2.25,2.6) {katabatic flow};
  \draw[thick, brown!70!black] (0.3,4.6) -- (3.5,4.3);
  \node[right, font=\footnotesize, brown!70!black] at (3.55,4.45) {2\textdegree\ slope};
  \draw[<->, thin] (0,-0.35) -- (8.9,-0.35);
  \node[below, font=\footnotesize] at (4.45,-0.35) {267\,m (89 cells)};
  \draw[<->, thin] (9.3,0) -- (9.3,5);
  \node[right, font=\footnotesize, rotate=90, anchor=south] at (9.9,2.5) {198\,m (66 cells)};
  \draw[<->, thin, red!80!black] (4.6,1.2) -- (6.6,1.2);
  \node[below, font=\footnotesize, red!80!black] at (5.6,1.15) {60\,m};
  \node[green!40!black, font=\footnotesize, align=center] at (2.3,0.7) {Vegetation\\(cat.~7)};
  \node[font=\footnotesize, align=center] at (7.7,0.7) {Residential\\(cat.~2)};
  \node[font=\footnotesize] at (0.5,0.3) {W};
  \node[font=\footnotesize] at (8.4,0.3) {E};
\end{tikzpicture}
\caption{Simulation domain (plan view). Vegetation (green, west) upwind of the 60\,m parcel (red outline) produces cold air that drains eastward towards the residential area (red) under pure katabatic forcing (2° downward terrain slope, no regional wind). The domain extends $66 \times 89$ cells at 3\,m resolution.}
\label{fig:domain}
\end{figure}
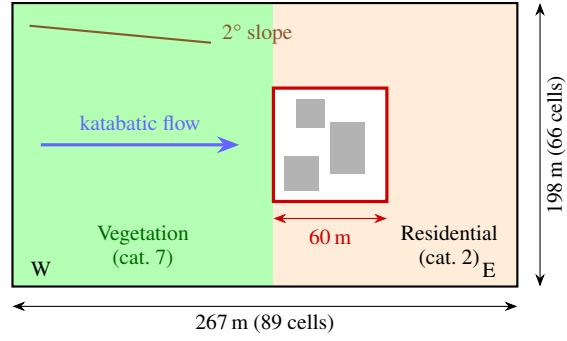

As a controlled proof-of-concept, we evaluate layouts on isolated 60\,m square parcels at 3\,m grid resolution (Fig.~\ref{fig:domain}).
The terrain features a continuous 2° downhill slope (west to east), providing gravitational forcing for pure katabatic cold air drainage ($v_{\text{regio}} = 0$).
Upwind of the parcel, land use is grassland (KLAM\_21 category~7, enabling cold air production); the parcel and downwind area use low-density residential (category~2).
Buffer zones extend the domain to $66 \times 89$ cells ($198\,\text{m} \times 267\,\text{m}$).
This simplified setup isolates the effect of building morphology on cold-air flux; the reported accuracy metrics therefore represent upper bounds in real urban contexts.

\section{Method: Three-Stage Surrogate Pipeline}
\label{sec:method}

\begin{figure}[t]
    \centering
    \includegraphics[width=\textwidth]{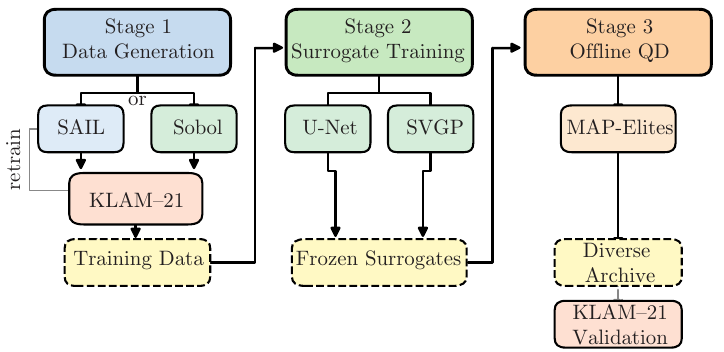}
    \caption{Three-stage experimental pipeline. Stage~1: Training data is generated either via \gls{sail} (physics-in-the-loop \gls{qd}) or via Sobol sampling with KLAM\_21 evaluation. Stage~2: A U-Net and/or \gls{svgp} surrogate is trained on the resulting data. Stage~3: Offline MAP-Elites uses the frozen surrogate to produce thousands of diverse layouts in minutes. A subset is validated with KLAM\_21.}
    \label{fig:pipeline}
\end{figure}

Our pipeline consists of three stages (Fig.~\ref{fig:pipeline}): (1)~generating training data via physics simulation, (2)~training surrogate models on the resulting data, and (3)~running offline MAP-Elites with the frozen surrogate. We describe each stage below.

\subsection{Stage 1: Data Generation via SAIL}
\label{sec:sail}

\gls{sail}~\cite{Gaier2018DataEfficient} combines MAP-Elites with a \gls{gp} surrogate model.
In each iteration, the \gls{gp} is trained on all solutions evaluated so far. An acquisition function (\gls{ucb}: $\alpha(x) = \mu(x) + \lambda \cdot \sigma(x)$) is used as a cheap proxy for fitness inside MAP-Elites.
After a fixed number of surrogate-only generations, the most promising solutions (acquisition map elites) are evaluated with the real KLAM\_21 simulator, and the \gls{gp} is retrained.

Critically, the \gls{sail} archive stores \emph{surrogate-predicted} fitness values.
For downstream use as supervised training data, we re-evaluate all archive elites with the true KLAM\_21 simulator to obtain ground-truth labels.
This is a one-time batch computation that runs in parallel on an \gls{hpc} cluster.

\paragraph{Training data comparison.}
We compare three data sources: (i)~\gls{sail} archives (re-evaluated with KLAM\_21 for ground-truth labels), (ii)~quasi-random Sobol sequences evaluated with KLAM\_21, and (iii)~a 50/50 combination.
\gls{ucb} acquisition within \gls{sail} drives exploration of uncertain regions, while MAP-Elites diversity pressure ensures broad feature-space coverage.
Whether this coverage advantage translates to better downstream surrogate accuracy -- and whether the answer depends on the surrogate architecture -- is a central question of this work.

\subsection{Stage 2: Surrogate Models}
\label{sec:surrogates}

We compare two surrogate architectures for use in Stage~3 (offline \gls{qd}).
Both are trained offline on KLAM\_21-evaluated data and are distinct from the online \gls{gp} retrained iteratively within \gls{sail} (Stage~1).

\paragraph{\Gls{svgp}.}
Following Hensman et al.~\cite{Hensman2015}, we train an \gls{svgp} with a Mat\'ern-2.5 kernel with \gls{ard} over 62 input dimensions (60-gene genome $+$ parcel width and height).
The model uses 2{,}000--5{,}000 inducing points, is implemented in GPyTorch~\cite{Gardner2018}, and predicts a scalar fitness value with calibrated uncertainty.
Training uses early stopping with a patience of 20 epochs, a learning-rate warmup over 10 epochs, and a batch size of 1{,}024.

\begin{figure}[t]
\centering
\includegraphics[width=\textwidth]{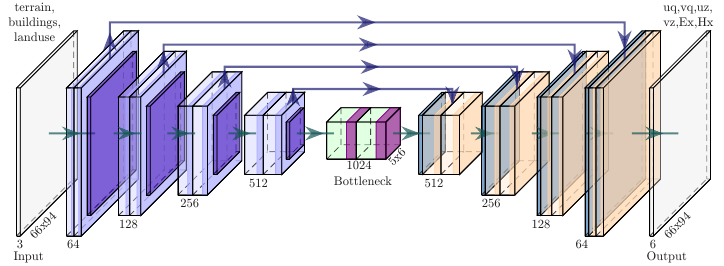}
\caption{U-Net architecture (depth 4, 64 base channels). Numbers underneath blocks indicate channel count; block heights are proportional to channels. Each encoder/decoder level applies two $3 \times 3$ convolutions with batch normalization and ReLU. Encoder levels downsample via max-pooling ($\downarrow$2); decoder levels upsample bilinearly ($\uparrow$2) and concatenate skip connections. A final $1 \times 1$ convolution projects to 6 output channels.}
\label{fig:unet}
\end{figure}

\paragraph{U-Net.}
A U-Net~\cite{Ronneberger2015} with 4 encoder/decoder stages and 64 base filters (Fig.~\ref{fig:unet}) takes 3-channel spatial input (terrain, building heights, land use) and predicts 6 output channels corresponding to KLAM\_21 output fields ($u_q, v_q, u_z, v_z, E_x, H_x$).
The scalar cold air flux objective (Eq.~\ref{eq:flux}) is then derived from the predicted spatial fields.
This spatial inductive bias -- predicting \emph{where} flow occurs, not just an aggregate scalar -- is a key advantage, as it constrains the model to learn physically plausible spatial patterns.

\subsection{Stage 3: Offline MAP-Elites}
\label{sec:offline}

In the final stage, MAP-Elites runs entirely with the frozen U-Net or \gls{svgp} surrogate -- no physics evaluations at all.
We test three configurations:
\textbf{(a)~U-Net only}: Fitness $= f_{\text{U-Net}}(x)$.
\textbf{(b)~\Gls{svgp} only}: Fitness $= \mu_{\text{GP}}(x)$, optionally with \gls{ucb}: $\mu(x) + \lambda \sigma(x)$.
\textbf{(c)~Hybrid}: Fitness $= f_{\text{U-Net}}(x) + \lambda \cdot \sigma_{\text{GP}}(x)$, combining U-Net accuracy with \gls{gp} uncertainty for exploration.
After optimization, a subset of archive elites is validated with the real KLAM\_21 simulator to assess the fidelity of the surrogate-predicted rankings.

\section{Experimental Setup}
\label{sec:setup}

\subsection{Data Generation}

We generated approximately 19{,}000 KLAM\_21 simulation runs across multiple parcel sizes.
For the \gls{sail}-based data, we ran \gls{sail} with real KLAM\_21 evaluations and then re-evaluated all archive elites with KLAM\_21 to obtain ground-truth fitness labels (the archives store only surrogate-predicted values).
For the random baseline, Sobol quasi-random sequences were evaluated directly with KLAM\_21.
\gls{svgp} experiments use the 62D genome-based representation; U-Net experiments use 3-channel spatial grids at native resolution.

\subsection{Experiments Overview}

Table~\ref{tab:experiments} summarizes the four experiments conducted.
All experiments use 3 replicates with different random seeds; we report mean $\pm$ standard deviation.

\begin{table}[t]
\centering
\caption{Overview of experiments, research questions, and key metrics.}
\label{tab:experiments}
\small
\resizebox{\textwidth}{!}{
\begin{tabular}{@{}clll@{}}
\toprule
\# & Research Question & Method & Key Metric \\
\midrule
1 & Training data strategy (RQ1) & Cross-domain SVGP evaluation & $R^2$ \\
2 & SVGP hyperparameters (RQ2) & Grid search: inducing points & $R^2$, calibration \\
3 & U-Net accuracy (RQ2) & Train/test on SAIL \& Sobol & $R^2$ per variable \\
4 & QD with surrogates (RQ3) & U-Net/SVGP/Hybrid MAP-Elites & Spearman $\rho$, QD ratio \\
\bottomrule
\end{tabular}}
\end{table}

\subsection{Evaluation Protocol}

For surrogate accuracy (Exp.~1--3), we use train/test splits and report $R^2$, RMSE, Spearman rank correlation, and 95\% confidence interval calibration (for \gls{svgp}).
\emph{Cross-domain evaluation} trains on one data source and tests on all three (optimized, Sobol, combined), revealing generalization failure modes.

For \gls{qd} validation (Exp.~4), we sample 100 elites from the surrogate-optimized archive and re-evaluate them with KLAM\_21.
We report: Spearman $\rho$ between surrogate-predicted and true fitness (ranking fidelity), $R^2$ between predictions and truth, and the \emph{\gls{qd} ratio}:
\begin{equation}
\label{eq:qdratio}
    \text{QD ratio} = \frac{\sum_{i \in \mathcal{A}} f_{\text{true}}(x_i)}{\sum_{i \in \mathcal{A}} f_{\text{surr}}(x_i)}
\end{equation}
where $\mathcal{A}$ is the set of validated archive elites, $f_{\text{true}}$ is the KLAM\_21-evaluated fitness, and $f_{\text{surr}}$ is the surrogate-predicted fitness.
Values near 1.0 indicate the surrogate neither over- nor under-estimates archive quality.
This metric is not standard in the \gls{qd} literature; it complements \gls{qd} score~\cite{Mouret2015} by measuring surrogate calibration rather than archive quality per se.

\section{Results}
\label{sec:results}

\subsection{Training Data Strategy (Exp.~1)\label{sec:res_training}}

We trained \gls{svgp} models on three data sources -- \gls{sail} archives (optimized), Sobol sequences (random), and a 50/50 combination -- and evaluated each model on all three test sets.
Table~\ref{tab:crossdomain} shows the average cross-domain $R^2$.

\begin{table}[t]
\centering
\caption{Cross-domain SVGP evaluation ($R^2$). Combined data generalizes best. Sobol-trained models fail catastrophically on optimized designs.}
\label{tab:crossdomain}
\small
\begin{tabular}{@{}lcccr@{}}
\toprule
Training data & $\to$ Optimized & $\to$ Sobol & $\to$ Combined & Avg.\ $R^2$ \\
\midrule
\textbf{Combined (50/50)} & \textbf{0.964} & \textbf{0.974} & \textbf{0.967} & \textbf{0.968} \\
Optimized (SAIL) & 0.964 & 0.889 & 0.929 & 0.927 \\
Random (Sobol) & $-0.000$ & 0.981 & 0.491 & 0.490 \\
\bottomrule
\end{tabular}
\end{table}

The key finding is twofold.
First, \textbf{combined training data} achieves the best average cross-domain $R^2$ of 0.968.
Second -- and more striking -- a model trained exclusively on Sobol data \textbf{fails catastrophically} on optimized designs ($R^2 \approx 0$), despite achieving excellent in-distribution accuracy ($R^2 = 0.981$).
Sobol-sampled designs concentrate in low-coverage, low-fitness regions of the search space and never expose the model to the high-fitness configurations that \gls{qd} algorithms will produce.

\begin{figure}[t]
    \centering
    \includegraphics[width=\textwidth]{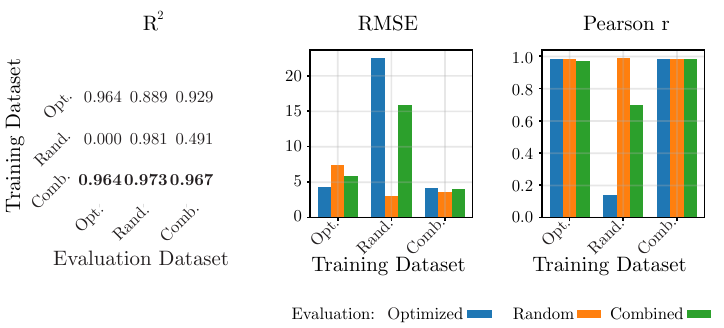}
    \caption{Cross-domain SVGP evaluation (Exp.~1). Each panel shows predicted vs.\ true flux for a model trained on one data source (columns) and tested on another (rows). Sobol-trained models fail on optimized test data, while combined-trained models generalize across all domains.}
    \label{fig:crossdomain}
\end{figure}

\gls{sail} archives, in contrast, span the full fitness landscape because \gls{ucb} acquisition drives exploration of uncertain regions while MAP-Elites diversity pressure covers the feature space.
The combined dataset merges the strengths of both: broad quasi-random coverage and targeted high-fitness representation.
Fig.~\ref{fig:crossdomain} visualizes the cross-domain evaluation matrix, showing the catastrophic failure pattern of Sobol-trained models.

\subsection{SVGP Hyperparameter Optimization (Exp.~2)\label{sec:res_hpo}}

\begin{figure}[t]
    \centering
    \includegraphics[width=\textwidth]{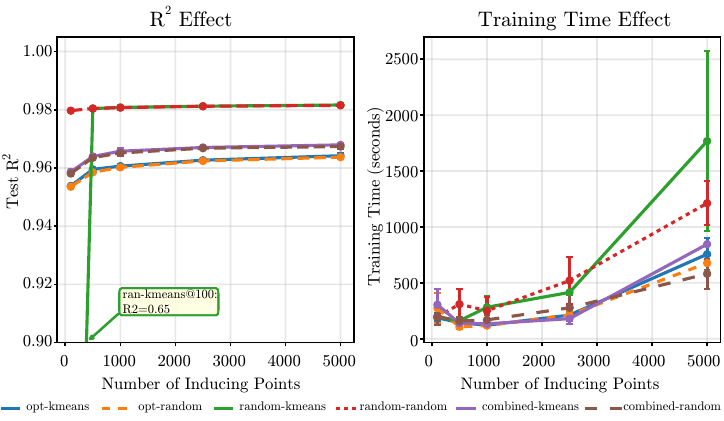}
    \caption{Effect of inducing point count on SVGP test $R^2$ (left) and training time (right). All configurations with $\geq$500 inducing points achieve $R^2 > 0.95$. The ran-km configuration at 100 points ($R^2 = 0.65$, off-chart) suffered from a single divergent replicate.}
    \label{fig:inducing}
\end{figure}

\textcolor{blue}{Figure~\ref{fig:inducing} shows} a grid search over inducing point count (100 to 5{,}000) and initialization method.
Accuracy improves monotonically with inducing points, saturating around 2{,}500--5{,}000 ($R^2 = 0.968$, calibration 95.6\%).
K-means initialization yields $<$0.1\% improvement over random subset selection.
Inference throughput ranges from 430{,}000 samples/s (100 pts) to 59{,}000 samples/s (5{,}000 pts).
Critically, hyperparameter optimization cannot compensate for poor training data: the best Sobol-trained \gls{svgp} still achieves only $R^2 \approx 0$ on optimized test data regardless of HPO configuration, confirming the primacy of training data selection (Sec.~\ref{sec:res_training}).

\subsection{U-Net Surrogate Accuracy (Exp.~3)\label{sec:res_unet}}

We trained U-Net models on both \gls{sail} and Sobol data with two loss functions (MSE and MSE\,+\,gradient), 3 seeds each (12 configurations total).
Table~\ref{tab:unet} shows that U-Net achieves $R^2 \geq 0.994$ in all configurations.
Per-variable $R^2$ ranges from 0.985 ($v_q$, cross-wind component at 2\,m) to 0.999 ($E_x$, cold air content); all six output variables exceed $R^2 = 0.98$.
While an $R^{2}$ approaching 1.0 might typically raise concerns of overfitting, it is important to note that the surrogate is learning a deterministic physics simulator on a highly constrained, simplified domain. The underlying mapping is exceptionally smooth and learnable, and this high accuracy reflects the controlled nature of our proof-of-concept rather than overfitting to noise.

\begin{table}[t]
\centering
\caption{U-Net accuracy by training configuration. Near-perfect accuracy regardless of data source or loss function. Mean $\pm$ std over 3 seeds.}
\label{tab:unet}
\small
\begin{tabular}{@{}llcc@{}}
\toprule
Data source & Loss & Overall $R^2$ & Overall MSE \\
\midrule
Sobol & MSE & $0.996 \pm 0.000$ & $0.004 \pm 0.000$ \\
Sobol & MSE+grad & $0.996 \pm 0.000$ & $0.004 \pm 0.000$ \\
SAIL & MSE & $0.994 \pm 0.000$ & $0.006 \pm 0.000$ \\
SAIL & MSE+grad & $0.994 \pm 0.000$ & $0.006 \pm 0.000$ \\
\bottomrule
\end{tabular}
\end{table}

\begin{figure}[t]
    \centering
    \includegraphics[width=\textwidth]{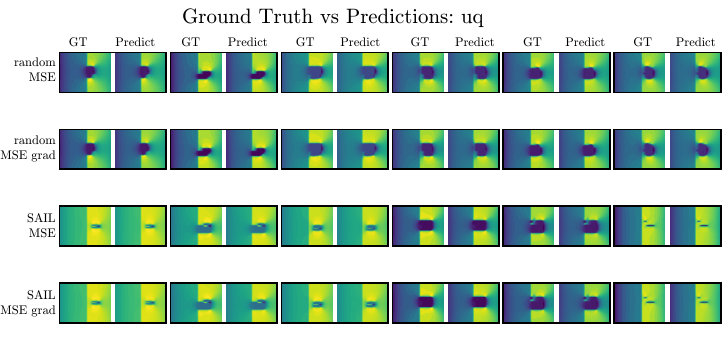}
    \caption{U-Net spatial predictions vs.\ KLAM\_21 ground truth for the $u_q$ wind velocity component (Exp.~3). Rows show different datasets (random, SAIL) and loss functions (MSE, MSE\_GRAD). Each sample shows the ground truth (GT) and the prediction (pred). The U-Net models capture fine-grained flow patterns around buildings with minimal error.}
    \label{fig:unet_predictions}
\end{figure}

The ground truth and predicted Figure~\ref{fig:unet_predictions} shows the predicted $u_q$ values for several samples for each training data set and loss function.
Unlike the \gls{svgp} (Sec.~\ref{sec:res_training}), \textbf{U-Net accuracy is effectively independent of training data source} -- Sobol and \gls{sail} data yield comparable results.
U-Net operates on spatial grids and learns the underlying physics mapping (input fields $\to$ output fields), rather than memorizing scalar input--output pairs from a particular region of genome space.
This data-source independence eliminates the need for the expensive \gls{sail} bootstrapping stage: trivially parallelizable Sobol samples suffice.

\subsection{QD with Offline Surrogates (Exp.~4)\label{sec:res_qd}}

\paragraph{\gls{svgp}-based MAP-Elites.}
Using the \gls{svgp} as sole fitness evaluator (5{,}000 generations, 9 configurations), archives grow to 2{,}300--5{,}500 elites but show \textbf{strongly negative validation $R^2$} ($-3.8$ to $-33.2$) and low Spearman $\rho$ (0.19--0.47) -- the \gls{gp} develops spurious optima that MAP-Elites exploits.
This motivates the U-Net-based approach.

\begin{figure}[t]
    \centering
    \includegraphics[width=\textwidth]{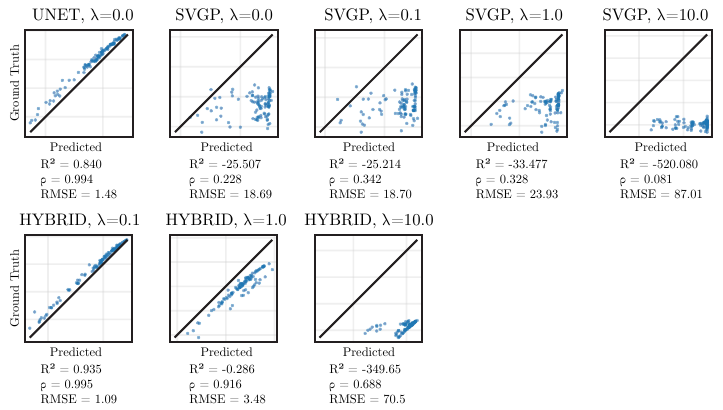}
    \caption{QD validation (Exp.~4): surrogate-predicted vs.\ KLAM\_21-validated fitness for 100 archive elites per configuration. U-Net shows tight clustering around the diagonal ($\rho = 0.994$); SVGP configurations exhibit severe prediction divergence due to exploitation of spurious optima. The hybrid surrogate model, which uses U-NET for prediction and SVGP for exploration, shows the most accurate performance when $\lambda$ is low, where SAIL is less explorative.}
    \label{fig:qd_validation}
\end{figure}

\paragraph{U-Net, \gls{svgp}, and Hybrid \gls{qd}.}
We systematically compare 8 configurations (U-Net only, \gls{svgp} only, \gls{svgp}+\gls{ucb} with $\lambda \in \{0.1, 1.0, 10.0\}$, and Hybrid with $\lambda \in \{0.1, 1.0, 10.0\}$), each with 3 seeds and 100 KLAM\_21-validated samples per archive.
Table~\ref{tab:qd} and Figure~\ref{fig:qd_validation} show the validated results.

\begin{table}[t]
\centering
\caption{QD validation: surrogate-predicted vs.\ KLAM\_21 ground truth (100 validated elites per archive). Mean $\pm$ std over 3 seeds. Tier A configurations are statistically equivalent.}
\label{tab:qd}
\small
\begin{tabular}{@{}llccc@{}}
\toprule
Tier & Configuration & Spearman $\rho$ & QD Ratio & Validation $R^2$ \\
\midrule
\multirow{2}{*}{A} & \textbf{U-Net ($\lambda = 0$)} & $0.994 \pm 0.001$ & $\mathbf{1.049 \pm 0.000}$ & $0.852 \pm 0.018$ \\
 & Hybrid ($\lambda = 0.1$) & $\mathbf{0.995 \pm 0.000}$ & $1.033 \pm 0.001$ & $0.924 \pm 0.008$ \\
\midrule
B & Hybrid ($\lambda = 1.0$) & $0.933 \pm 0.017$ & $0.890 \pm 0.009$ & $-0.29 \pm 0.24$ \\
\midrule
C & Hybrid ($\lambda = 10.0$) & $0.731 \pm 0.033$ & $0.294 \pm 0.005$ & $-323 \pm 50$ \\
\midrule
\multirow{2}{*}{D} & SVGP ($\lambda = 0$) & $0.200 \pm 0.165$ & $0.611 \pm 0.016$ & $-21 \pm 3$ \\
 & SVGP ($\lambda = 0.1$) & $0.348 \pm 0.059$ & $0.608 \pm 0.002$ & $-23 \pm 4$ \\
\midrule
\multirow{2}{*}{E} & SVGP ($\lambda = 1.0$) & $0.404 \pm 0.070$ & $0.548 \pm 0.011$ & $-33 \pm 1$ \\
 & SVGP ($\lambda = 10.0$) & $0.048 \pm 0.037$ & $0.238 \pm 0.003$ & $-585 \pm 52$ \\
\bottomrule
\end{tabular}
\end{table}

Although the Hybrid model ($\lambda = 0.1$) achieves a marginally better Spearman $\rho$ (0.995) and QD ratio (1.033) than the pure U-Net, we select the pure U-Net ($\lambda = 0$) as the recommended configuration. The negligible performance gain of the hybrid approach does not justify the added computational overhead and system complexity required to maintain and evaluate an SVGP alongside the U-Net during practical deployment.

The results yield three conclusions:
\textbf{(1)~U-Net alone is sufficient.} Pure U-Net ($\rho = 0.994$) and the best hybrid ($\rho = 0.995$) are statistically equivalent; the \gls{qd} ratio of 1.049 indicates the U-Net is marginally conservative, which is desirable.
\textbf{(2)~\Gls{svgp} alone is inadequate.} All \gls{svgp}-only configurations achieve $\rho < 0.41$ with deeply negative validation $R^2$, confirming that genome-based scalar surrogates are ill-suited for offline \gls{qd} in spatially structured domains.
\textbf{(3)~High \gls{ucb} $\lambda$ is counterproductive.} Uncertainty-driven exploration in a frozen-model setting pushes solutions toward unreliable regions.

\subsection{Practical Deployment}
\label{sec:res_deployment}

Performance optimization (Numba JIT feature computation, vectorized domain grid construction, batch encoding) reduces the full evaluation pipeline from 823\,ms to 187\,ms per generation (batch of 1{,}024).
Raw surrogate throughput is $\sim$465 samples/s for U-Net and $\sim$2{,}857 samples/s for \gls{svgp}; the \gls{svgp}'s unique advantage is calibrated predictive uncertainty.
A complete \gls{qd} run (10{,}000 generations, $\sim$80{,}000 unique evaluations) completes in 31--53 minutes; a practical ``coffee break'' configuration (1{,}000 generations) runs in 3--8 minutes.
The $>12,000\times$ per-evaluation speedup is amortized after the one-time cost of $\sim$19,000 training simulations ($\sim$130 hours, trivially parallelizable on HPC). Training the U-Net surrogate itself takes approximately 20 minutes on a single NVIDIA A100 GPU.
The pipeline is deployed in the open-source OpenSKIZZE web application, integrating German geodata standards with a 6-step workflow: parcel selection, constraint definition, \gls{qd} optimization, archive exploration, cluster analysis, and 3D comparison with flow overlays.
\section{Discussion}
\label{sec:discussion}

\paragraph{Why U-Net succeeds where \gls{svgp} fails: a representation issue.}
The comparison between \gls{svgp} and U-Net in this work is not a comparison of model classes (\gls{gp} vs.\ neural network) but of \emph{representation paradigms}: genome-based scalar prediction (62D input $\to$ scalar fitness) vs.\ spatial field-to-field prediction ($66 \times 89 \times 3$ input $\to$ $66 \times 89 \times 6$ output).
The U-Net leverages convolutional spatial structure to produce physically coherent flow fields; the \gls{svgp}, operating on an opaque genome vector, lacks any spatial inductive bias.
Using the raw spatial input for a \gls{gp} is computationally infeasible: flattening the 3-channel input grid yields $\sim$17{,}600 dimensions, far beyond the scalability of \gls{gp} inference.
A neural network predicting scalar fitness from genomes could plausibly exhibit the same spurious-optima failure as the \gls{svgp} -- it is the spatial representation, not the model class, that prevents the U-Net from producing arbitrary fitness landscapes.
We hypothesize that this spatial inductive bias constrains predictions to be physically plausible, reducing the risk of spurious optima in offline \gls{qd}.
Verifying this hypothesis on more complex domains with out-of-distribution inputs is an important direction for future work.
More broadly, this result suggests that \textbf{spatial surrogates should be preferred over scalar genome-based surrogates for offline \gls{qd} optimization} whenever the underlying physics has spatial structure.

\paragraph{The bootstrap cost.}
For U-Net, Sobol data suffices (Sec.~\ref{sec:res_unet}), so no \gls{sail} run is needed.
For \gls{svgp}, \gls{sail} archives remain essential; combined training data is recommended when only scalar surrogates are available.

\paragraph{Prototype limitations.}
Our proof-of-concept uses simplified boundary conditions: isolated 60\,m parcels, continuous 2° terrain slope, pure katabatic flow.
The reported $R^2$ values are upper bounds; sensitivity analysis shows that boundary conditions significantly affect absolute flux magnitudes. Heterogeneous land use and variable wind directions will also degrade model performance.
We used 3 replicates per configuration; while variance is consistently low (Table~\ref{tab:qd}), future work should use $\geq$10 replicates.
Ongoing work trains multi-scale U-Nets on $\sim$3\,km $\times$ 3\,km real urban domains with variable wind directions.

\paragraph{Adaptive Learning and Uncertainty Quantification.} We chose a purely offline phase because interleaving deep spatial model retraining during optimization introduces prohibitive computational bottlenecks compared to lightweight GPs. However, adaptive learning remains a compelling path for hardware-accelerated workflows. Additionally, while the SVGP natively provided uncertainty quantification (UQ), it failed during optimization. Integrating UQ directly into the spatial surrogate via Bayesian Neural Networks or deep ensembles is a critical step for ensuring reliability in regulatory urban planning.

\paragraph{Generalizability.}
The pipeline (data generation $\to$ deep surrogate $\to$ offline \gls{qd}) is domain-agnostic, requiring only an expensive black-box simulator with spatial I/O and a \gls{qd} formulation with meaningful features.
Potential applications include aerodynamic shape optimization, structural design, and other urban planning objectives.

\section{Conclusion}
\label{sec:conclusion}

We presented a surrogate-assisted \gls{qd} pipeline to optimize urban building layouts for cold-air ventilation.
A U-Net trained on Sobol-sampled physics simulations predicts KLAM\_21 cold air flow fields at $R^2 = 0.996$, enabling offline MAP-Elites that produce diverse building layouts with physics-validated ranking fidelity of $\rho = 0.994$ -- a $>$12{,}000$\times$ per-evaluation speedup over direct simulation (amortized from the second \gls{qd} run onward, after one-time training data generation).
We note that the very high surrogate accuracy is partly attributable to the simplified, single-parcel layouts considered here; real urban morphologies with irregular geometries and heterogeneous land use are likely to present a harder learning problem.
Our systematic evaluation yielded actionable insights: combined \gls{qd}-archive and Sobol training data generalizes best for scalar surrogates, but spatial surrogates achieve equivalent accuracy with Sobol data alone; deep spatial surrogates outperform genome-based scalar surrogates for offline \gls{qd} due to spatial inductive biases; and U-Net accuracy is robust to training data source, eliminating the need for expensive bootstrapping.

Future work focuses on three directions: (i)~multi-scale U-Nets handling variable parcel sizes and real urban contexts -- we are currently scaling to 3\,km\,$\times$\,3\,km domains with heterogeneous building stock, (ii)~conditional generative models to replace the parametric encoding with direct, constraint-aware layout generation, and (iii) extension to multi-objective optimization, utilizing multiple objectives and modeling correlations to balance cold-air ventilation with other critical urban metrics such as heat stress, thermal comfort, and solar potential.

The pipeline is available as the open-source OpenSKIZZE tool. Application and data are published at \url{https://github.com/alexander-hagg/openskizze} -- experiments at \url{https://github.com/alexander-hagg/openskizze-ppsn2026.git}.

\subsection{Acknowledgements}
This work was funded by the German Federal Environmental Foundation (Deutsche Bundesstiftung Umwelt, DBU), grant no.\ 39022/01.
We thank the \gls{dwd} for providing the KLAM\_21 model and technical support.

\subsection{Competing Interests}
The authors have no competing interests to declare.

\bibliographystyle{apalike}
\bibliography{bib}

\end{document}